\documentclass{article}

\usepackage{arxiv}
\usepackage{amsfonts}

\usepackage[utf8]{inputenc} 
\usepackage[T1]{fontenc}    
\usepackage{hyperref}       
\hypersetup{colorlinks=true, linkcolor=blue, filecolor=blue,  urlcolor=blue, citecolor=blue }
\usepackage{url}            
\usepackage{booktabs}       
\usepackage{amsfonts}       
\usepackage{nicefrac}       
\usepackage{microtype}      
\usepackage{lipsum}		
\usepackage{graphicx}
\usepackage{natbib}
\usepackage{doi}
\usepackage{float}
\usepackage{longtable}
\usepackage{amsmath}
\usepackage{csquotes}

\newcommand*{\escape}[1]{\texttt{\textbackslash#1}}

\title{Comparing Variation in Tokenizer Outputs Using a Series of Problematic and Challenging Biomedical Sentences}

\author{ 
    \href{https://orcid.org/0000-0002-5429-5233}{\includegraphics[scale=0.06]{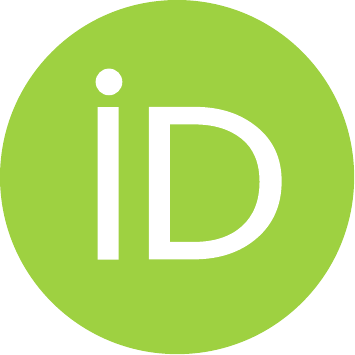}\hspace{1mm}Christopher Meaney}\\
	Department of Family and Community Medicine\\
	University of Toronto\\
	Toronto, Ontario, Canada \\
	\texttt{christopher.meaney@utoronto.ca}
	\And
	\href{https://orcid.org/0000-0001-9283-8764}{\includegraphics[scale=0.06]{orcid.pdf}\hspace{1mm}Therese A Stukel}\\
	ICES\\
	Toronto, Ontario, Canada \\
	\texttt{therese.stukel@ices.on.ca}
	\And
	\href{https://orcid.org/0000-0003-3337-233X}{\includegraphics[scale=0.06]{orcid.pdf}\hspace{1mm}Peter C Austin}\\
        ICES\\
 	Toronto, Ontario, Canada \\
	\texttt{peter.austin@ices.on.ca}
	\And
	\href{https://orcid.org/0000-0001-9055-4709}{\includegraphics[scale=0.06]{orcid.pdf}\hspace{1mm}Michael Escobar}\\
	Dalla Lana School of Public Health\\
	University of Toronto\\
	Toronto, Ontario, Canada \\
	\texttt{m.escobar@utoronto.ca}	
	}

\renewcommand{\headeright}{}
\renewcommand{\undertitle}{Technical Report}

\hypersetup{
pdftitle={Comparing Variation in Tokenizer Outputs Using a Series of Problematic and Challenging Biomedical Sentences},
pdfsubject={cs.CL},
pdfauthor={Christopher Meaney, Therese A Stukel, Peter C Austin, Michael Escobar},
pdfkeywords={Biomedical text data, Text mining, Tokenizers},
}

\begin{document}
\maketitle

\begin{abstract}

\textbf{Background \& Objective:} 
Biomedical text data are increasingly available for research. Tokenization is an initial step in many biomedical text mining pipelines. Tokenization is the process of parsing an input biomedical sentence (represented as a digital character sequence) into a discrete set of word/token symbols, which convey focused semantic/syntactic meaning. The objective of this study is to explore variation in tokenizer outputs when applied across a series of problematic and challenging biomedical sentences.

\textbf{Method:} 
\citet{diaz2015analysis} introduce 24 challenging example biomedical sentences for comparing tokenizer performance. In this study, we descriptively explore variation in outputs of eight tokenizers applied to each example biomedical sentence. The tokenizers compared in this study are the NLTK white space tokenizer, the NLTK Penn Tree Bank tokenizer, Spacy and SciSpacy tokenizers, Stanza/Stanza-Craft tokenizers, the UDPipe tokenizer, and R-tokenizers.

\textbf{Results:} 
For many examples, tokenizers performed similarly effectively; however, for certain examples, there were meaningful variation in returned outputs. The white space tokenizer often performed differently than other tokenizers (appending punctuation suffixes to tokens/words). We observed performance similarities for tokenizers implementing rule-based systems (e.g. pattern matching and regular expressions) and tokenizers implementing neural architectures for token classification. Oftentimes, the challenging tokens resulting in the greatest variation in outputs, are those words which convey substantive and focused biomedical/clinical meaning (e.g. x-ray, IL-10, TCR/CD3, CD4+ CD8+, and (Ca2+)-regulated).

\textbf{Conclusion:} 
When state-of-the-art, open-source tokenizers from Python and R were applied to a series of challenging biomedical example sentences, we observed subtle variation in the returned outputs. Data scientists engaging with text mining should be familiar with the landscape of tokenizers available for their research problem, and how the choice of tokenizer impacts downstream inferences.  

\end{abstract}


\section{Introduction}

With the increasing digitization of human communication, we are collecting growing volumes of text data, which can be used for research purposes \citep{gentzkow2019text}. Models and algorithms are required to extract meaningful insights from these voluminous collections of text data. 

Statistical semantic models leverage patterns of human word usage (e.g. word occurrence statistics, word collocation frequencies, etc.) to extract meaning from large document collections \citep{turney2010frequency} \citep{manning1999foundations}. Vector space models, such as document term matrices (DTM) or term co-occurrence matrices (TCM), use sparse high-dimensional arrays to represent important features of large document collections. Elements of these arrays are count random count variables, denoting the number of times a linguistic unit occurred under some context. For example, an element (d,v) for (d=1…D, v=1…V) of the DTM counts the number of times a particular word/token (v) in the corpus, occurred in a document/context (d). Similarly, an element of the TCM counts the number of times a word in the corpus (j $\in$ 1…V) occurred within a certain context/window-width from every other word in the corpus (i $\in$ 1…V). These mathematical arrays encode salient information about the corpus and can be integrated into supervised/unsupervised learning pipelines. For example:

\begin{itemize}
    \item Supervised regression/classification on DTM features.
    \item Unsupervised document clustering using DTM features (e.g. multinomial mixture models).
    \item Unsupervised topic modelling on DTM features (e.g. LSA, LDA, NMF, etc.).
    \item Unsupervised word-embeddings using TCM features (e.g. GLoVe).
\end{itemize}

The above models rely on a precise operational definition of a “word” (v=1…V) for construction. Input documents are represented as digital character sequences. Tokenization algorithms are computational tools which break/slice/parse a digital character sequence into words, numbers, punctuations, and other symbols \citep{webster1992tokenization} \citep{manning1999foundations} \citep{he2006comparison}. Tokenizers effectively define the cardinality of the vocabulary/dictionary (v=1…V) of “words”, or more precisely “tokens” encountered in a text mining study. Variation in tokenizer output may result in different elements being included in the set of v=1…V unique words/tokens, and occurrence/co-occurrence counts being distorted. Hence, effectively understanding the performance (strengths, weaknesses, characteristics) of a tokenization algorithm is crucial to developing high performing statistical NLP (natural language processing) systems. 

Biomedical and clinical texts are especially challenging to tokenize, as they are often characterized by ungrammatical documents (e.g. spelling errors, abbreviations, author specific reporting styles, etc.); terms containing a mix of letters and digits, dashes, slashes, periods; and technical scientific/clinical language (e.g. chemical compounds, genetic information, diseases, etc.) \citep{jiang2007empirical} \citep{diaz2015analysis}. Because of these inherent complexities, we hypothesize that different tokenizers will vary in their returned outputs when applied to challenging biomedical sentences. The objective of this study is to review and compare several modern, open-source, tokenization algorithms available in the Python and R programming languages when applied against several challenging biomedical sentences.

\section{Methods}

\subsection{Study Purpose, Design and Descriptive Evaluation}

The objective of this study is to review and compare several modern, open-source, tokenization algorithms available in the Python and R programming languages. We validate and extend the illustrative evaluation of \citet{diaz2015analysis}, applying various tokenizers against potentially problematic/challenging cases, and inspecting outputs of the returned tokenizers. Diaz et al outline twenty-four examples of problematic and challenging sentences for a biomedical tokenizer to effectively convert to tokens. \citet{diaz2015analysis} descriptively explore variation in returned outputs across twelve tokenizers, when applied across the twenty-four biomedical sentences. They posit that the following types of biomedical words/entities would be difficult to tokenize and devise a series of illustrative examples to demonstrate this claim. 

\begin{itemize}
    \item Hyphenated compound words
    \item Words with letters/slashes
    \item Words with letters and apostrophes
    \item Words with letters and brackets
    \item Abbreviations in capital letters and acronyms
    \item Words with letters and periods
    \item Words with letters and numbers
    \item Words with numbers and one type of punctuation
    \item Numeration
    \item Hypertext markup
    \item URLs
    \item DNA Sequences
    \item Temporal Expressions
    \item Chemical Substances
\end{itemize}

Each of the twenty-four biomedical sentences/examples is processed into a variable length “bag of words”. These twenty-four seemingly simple sentences effectively illustrate and improve understanding of how tokenizers perform, and how/why they vary with respect to difficult word/entity types. This can help the analyst better understand the pros/cons of particular tokenizers, when applied across different domains (e.g. diverse biomedical and clinical document collections). 

\subsection{Tokenization Algorithms}

The eight tokenization algorithms being compared are available in Python and R. The tokenizers are open source, and easily installed across most operating systems. Details regarding the tokenization algorithms are given in Table 1.

Most tokenization algorithms rely on the sequential application of pattern matching and regular expressions to parse and normalize input biomedical sentences (represented as digital character sequences). In English language biomedical texts, it is common to initiate a sequential tokenization pipeline using a white-space tokenizer (i.e. split on  \escape{s}, \escape{n}, \escape{f}, \escape{t} character-types). Next, pattern matching rules are employed to normalize tokens: handling prefixes, suffixes, and infixes; matching particular tokens (e.g. stop-word lists); and handling of other special cases. Case-folding is possible when tokenizing (i.e. lower/upper/title-case conversion). Certain text pre-processing pipelines may also stem or lemmatize tokens to further normalize returned token sets.    
  
Few algorithms treat tokenization as a character tagging problem and use light-weight recurrent neural network architectures to identify tokens from within sentences. For example, biomedical sentences are represented as digital character sequences. Each individual digital character in the sequence is associated with a categorical tag/label denoting whether it is located at the beginning (B), inside (I), end (E), or outside (O) a token. A character level recurrent neural network is then used to predict categorical labels across new input sentences, effectively defining boundary spans for each token in the character sequence.  

\subsection{Evaluation of Tokenization Algorithms}

We descriptively evaluate variation in the performance of the following eight tokenizers: 1) the NLTK white space tokenizer, 2) the NLTK Penn Tree Bank tokenizer, 3) the spacy tokenizer, 4) the scispacy tokenizer, 5) the stanza tokenizer, 6) the stanza-craft tokenizer, 7) R-tokenizer, and 8) the UDPIPE tokenizer. We apply each of the eight tokenizers, to the twenty-four challenging biomedical sentences introduced in \citet{diaz2015analysis}. For each sentence (represented as a digital character sequence) we report the unique bags-of-tokens returned by the tokenization algorithms. We report the total number of tokens returned by each algorithm, and the total number of unique tokens. Aggregating over the twenty-four sentences, we investigate the cardinality of vocabulary implied by each tokenization algorithm. We use the Jaccard index to investigate agreement across each of the sets of returned tokens, and characterize similarity and differences in tokenizer performance. Given two sets A and B, the Jaccard index is defined as the cardinality of the intersection of A and B, divided by the cardinality of the union of A and B.

\begin{table}[H]
	\caption{Tokenizer name, language (Python/R), documentation URLs, and citations.  
Programming}
	\centering
    \scalebox{0.45}{
	\begin{tabular}{llllll}
		\toprule
 
Language & Tokenizer Package Name & Package URL & URL for Tokenizer Documentation & Class of Tokenizer & Citation \\

            \midrule
            
Python & nltk-space	& \url{https://www.nltk.org/} & \url{https://www.nltk.org/api/nltk.tokenize.simple.html} &
Regular Expression	& \citep{bird2009natural} \\
            \midrule
        & nltk-tb &	\url{https://www.nltk.org/} &
\url{https://www.nltk.org/api/nltk.tokenize.treebank.html} &
Regular Expression & \citep{bird2009natural} \\
            \midrule
        & spacy	& \url{https://spacy.io/} &
\url{https://spacy.io/api/tokenizer} &
Regular Expression & \citep{honnibal2017spacy} \\
	    \midrule
        & scispacy & \url{https://allenai.github.io/scispacy/} &
\url{https://github.com/allenai/scispacy/blob/main/scispacy/custom\_tokenizer.py} &
Regular Expression & \citep{neumann2019scispacy} \\
            \midrule
	& stanza & \url{https://stanfordnlp.github.io/stanza/} &
\url{https://stanfordnlp.github.io/stanza/tokenize.html} &
Recurrent Neural Network & \citep{qi2020stanza} \\
            \midrule
	& stanza-craft & \url{https://stanfordnlp.github.io/stanza/biomed\_model\_performance.html} &
\url{https://stanfordnlp.github.io/stanza/tokenize.html} &
Recurrent Neural Network & \citep{zhang2021biomedical} \\
            \midrule
R	& tokenizers & \url{https://lincolnmullen.com/software/tokenizers/} &
\url{https://github.com/ropensci/tokenizers/blob/master/R/basic-tokenizers.R} &
Regular Expression	& \citep{a2018fast} \\
            \midrule
	& udpipe & \url{https://bnosac.github.io/udpipe/en/index.html} &
\url{https://github.com/bnosac/udpipe/blob/master/R/udpipe\_parse.R} &
Gated Recurrent Unit & \citep{straka2017tokenizing} \\
        \bottomrule
	\end{tabular}
	}
	\label{tab:table1}
\end{table}

\section{Results}

\subsection{Evaluation of Tokenizers Applied to Several Challenging Biomedical Sentences}

In sections 3.1.1-3.1.14 we present the outputs of each of the eight tokenizers when applied to the twenty-four problematic and challenging biomedical sentences outlined in \citet{diaz2015analysis}. For each example we report the initial input sentence, and all uniquely returned outputs of the tokenizers (separating tokens using white space). 

\subsubsection{Hyphenated compound words}

\begin{displayquote}
\textbf{Example 1: “Normal chest x-ray.”}
\end{displayquote}

\begin{itemize}
    \item (NLTK-space)
    \begin{itemize}
        \item Normal chest x-ray.
    \end{itemize}
    \item (NLTK-tb, spacy, Stanza, StanzaCraft, Tokenizers) 
    \begin{itemize}
        \item Normal chest x-ray .
    \end{itemize}
    \item (SciSpacy, UDPIPE)
    \begin{itemize}
        \item Normal chest x - ray .
    \end{itemize}
\end{itemize}

\begin{displayquote}
\textbf{Example 2: “2-year 2-month old female with pneumonia.”}
\end{displayquote}

\begin{itemize}
    \item (NLTK-space)
    \begin{itemize}
        \item 2-year 2-month old female with pneumonia.
    \end{itemize}
    \item (NLTK-tb, spacy, Tokenizers) 
    \begin{itemize}
        \item 2-year 2-month old female with pneumonia . 
    \end{itemize}
    \item (SciSpacy, Stanza, StanzaCraft, UDPIPE)
    \begin{itemize}
        \item 2 - year 2 - month old female with pneumonia .
    \end{itemize}
\end{itemize}

\begin{displayquote}
\textbf{Example 3: “This may occur through the ability of IL-10 to induce expression of the gene.”}
\end{displayquote}

\begin{itemize}
    \item (NLTK-space)
    \begin{itemize}
        \item This may occur through the ability of IL-10 to induce expression of the gene.
    \end{itemize}
    \item (NLTK-tb, spacy, SciSpacy, Tokenizers)
    \begin{itemize}
        \item This may occur through the ability of IL-10 to induce expression of the gene . 
    \end{itemize}
    \item (Stanza, StanzaCraft, UDPIPE)
    \begin{itemize}
        \item This may occur through the ability of IL - 10 to induce expression of the gene .
    \end{itemize}
\end{itemize}

\subsubsection{Words with letters/slashes}

\begin{displayquote}
\textbf{Example 4: “The maximal effect is observed at the IL-10 concentration of 20 U/ml.”}
\end{displayquote}

\begin{itemize}
    \item (NLTK-space)
    \begin{itemize}
        \item The maximal effect is observed at the IL-10 concentration of 20 U/ml. 
    \end{itemize}
    \item (NLTK-tb, spacy, Tokenizers)
    \begin{itemize}
        \item The maximal effect is observed at the IL-10 concentration of 20 U/ml .  
    \end{itemize}
    \item (SciSpacy)
    \begin{itemize}
        \item The maximal effect is observed at the IL-10 concentration of 20 U / ml .
    \end{itemize}
    \item (Stanza)
    \begin{itemize}
        \item The maximal effect is observed at the IL -10 concentration of 20 U / ml .
    \end{itemize}
    \item (StanzaCraft)
    \begin{itemize}
        \item The maximal effect is observed at the IL - 10 concentration of 20 U/ml . 
    \end{itemize}
    \item (UDPIPE)
    \begin{itemize}
        \item The maximal effect is observed at the IL - 10 concentration of 20 U / ml . 
    \end{itemize}
\end{itemize}

\begin{displayquote}
\textbf{Example 5: “These results indicate that within the TCR/CD3 signal transduction pathway both PKC and calcineurin are required for the effective activation of the IKK complex and NF-kappaB in T lymphocytes.”}
\end{displayquote}

\begin{itemize}
    \item (NLTK-space)
    \begin{itemize}
        \item These results indicate that within the TCR/CD3 signal transduction pathway both PKC and calcineurin are required for the effective activation of the IKK complex and NF-kappaB in T lymphocytes.
    \end{itemize}
    \item (NLTK-tb, spacy, Tokenizers)
    \begin{itemize}
        \item These results indicate that within the TCR/CD3 signal transduction pathway both PKC and calcineurin are required for the effective activation of the IKK complex and NF-kappaB in T lymphocytes . 
    \end{itemize}
    \item (SciSpacy, Stanza, StanzaCraft)
    \begin{itemize}
        \item These results indicate that within the TCR / CD3 signal transduction pathway both PKC and calcineurin are required for the effective activation of the IKK complex and NF - kappaB in T lymphocytes .
    \end{itemize}
    \item (UDPIPE)
    \begin{itemize}
        \item These results indicate that within the TCR / CD 3 signal transduction pathway both PKC and calcineurin are required for the effective activation of the IKK complex and NF - kappaB in T lymphocytes . 
    \end{itemize}
\end{itemize}

\subsubsection{Words with letters and apostrophes}

\begin{displayquote}
\textbf{Example 6: “The false positive rate of our predictor was estimated by the method of D'Haeseleer and Church 1855 and used to compare it to other prediction datasets.”}
\end{displayquote}

 \begin{itemize}
    \item (NLTK-space)
    \begin{itemize}
        \item The false positive rate of our predictor was estimated by the method of D'Haeseleer and Church 1855 and used to compare it to other prediction datasets.
    \end{itemize}
    \item (NLTK-tb, spacy, SciSpacy, Stanza, StanzaCraft, Tokenizers)
    \begin{itemize}
        \item The false positive rate of our predictor was estimated by the method of D'Haeseleer and Church 1855 and used to compare it to other prediction datasets . 
    \end{itemize}
    \item (UDPIPE)
    \begin{itemize}
        \item The false positive rate of our predictor was estimated by the method of D' Haeseleer and Church 1855 and used to compare it to other prediction datasets . 
    \end{itemize}
\end{itemize}

\begin{displayquote}
\textbf{Example 7: “Small, scarred right kidney, below more than 2 standard deviations in size for patient's age.”}
\end{displayquote}

 \begin{itemize}
    \item (NLTK-space)
    \begin{itemize}
        \item Small, scarred right kidney, below more than 2 standard deviations in size for patient's age.
    \end{itemize}
    \item (NLTK-tb, spacy, SciSpacy, Stanza, StanzaCraft, Tokenizers, UDPIPE)
    \begin{itemize}
        \item Small , scarred right kidney , below more than 2 standard deviations in size for patient 's age . 
    \end{itemize}
\end{itemize}

\subsubsection{Words with letters and brackets}

\begin{displayquote}
\textbf{Example 8: “Of these, Diap1 has been most extensively characterized; it can block cell death caused by the ectopic expression of reaper, hid, and grim (reviewed in [26]).”}
\end{displayquote}

 \begin{itemize}
    \item (NLTK-space)
    \begin{itemize}
        \item Of these, Diap1 has been most extensively characterized  it can block cell death caused by the ectopic expression of reaper, hid, and grim (reviewed in [26]).
    \end{itemize}
    \item (NLTK-tb, spacy, SciSpacy, Stanza, StanzaCraft, Tokenizers, UDPIPE)
    \begin{itemize}
        \item Of these , Diap1 has been most extensively characterized   it can block cell death caused by the ectopic expression of reaper , hid , and grim ( reviewed in [ 26 ] ) .
    \end{itemize}
\end{itemize}

\subsubsection{Abbreviations in capital letters and acronyms}

\begin{displayquote}
\textbf{Example 9: “Mutants in Toll signaling pathway were obtained from Dr. S. Govind: cactE8, cactIIIG, and cactD13 mutations in the cact gene on Chromosome II.”}
\end{displayquote}

 \begin{itemize}
    \item (NLTK-space)
    \begin{itemize}
        \item Mutants in Toll signaling pathway were obtained from Dr. S. Govind: cactE8, cactIIIG, and cactD13 mutations in the cact gene on Chromosome II.
    \end{itemize}
    \item (NLTK-tb, spacy, SciSpacy, Stanza, StanzaCraft, Tokenizers)
    \begin{itemize}
        \item Mutants in Toll signaling pathway were obtained from Dr. S. Govind : cactE8 , cactIIIG , and cactD13 mutations in the cact gene on Chromosome II .
    \end{itemize}
    \item (UDPIPE)
    \begin{itemize}
        \item Mutants in Toll signaling pathway were obtained from Dr. S. Govind : cactE8 , cactIIIG , and cactD 13 mutations in the cact gene on Chromosome II .
    \end{itemize}
\end{itemize}

\begin{displayquote}
\textbf{Example 10: “The transcripts were detected in all the CD4- CD8-, CD4+ CD8+, CD4+ CD8-, and CD4- CD8+ cell populations.”}
\end{displayquote}

\begin{itemize}
    \item (NLTK-space)
    \begin{itemize}
        \item The transcripts were detected in all the CD4- CD8-, CD4+ CD8+, CD4+ CD8-, and CD4- CD8+ cell populations.
    \end{itemize}
    \item (NLTK-tb, Tokenizers)
    \begin{itemize}
        \item The transcripts were detected in all the CD4- CD8- , CD4+ CD8+ , CD4+ CD8- , and CD4- CD8+ cell populations . 
    \end{itemize}
    \item (spacy, SciSpacy)
    \begin{itemize}
        \item The transcripts were detected in all the CD4- CD8- , CD4 + CD8 + , CD4 + CD8- , and CD4- CD8 + cell populations .
    \end{itemize}
    \item (Stanza)
    \begin{itemize}
        \item he transcripts were detected in all the CD4 - CD8 - , CD4 + CD8 + , CD4 + CD8 - , and CD4 - CD8 + cell populations .
    \end{itemize}
    \item (StanzaCraft)
    \begin{itemize}
        \item The transcripts were detected in all the CD4 - CD8 - , CD4 + CD8 + , CD4 + CD8 -, and CD4- CD8+ cell populations . 
    \end{itemize}
    \item (UDPIPE)
    \begin{itemize}
        \item The transcripts were detected in all the CD4 - CD8 - , CD4 + CD8 + , CD4 + CD8 -, and CD4 - CD8 + cell populations .
    \end{itemize}
\end{itemize}

\subsubsection{Words with letters and periods}

\begin{displayquote}
\textbf{Example 11: “Two stop codons of an iORF (i.e. the inframe and C-terminal stops) can be any combination of canonical stop codons (TAA, TAG, TGA).”}
\end{displayquote}

\begin{itemize}
    \item (NLTK-space)
    \begin{itemize}
        \item Two stop codons of an iORF (i.e. the inframe and C-terminal stops) can be any combination of canonical stop codons (TAA, TAG, TGA).
    \end{itemize}
    \item (NLTK-tb, spacy, Stanza, StanzaCraft, Tokenizers)
    \begin{itemize}
        \item Two stop codons of an iORF ( i.e. the inframe and C-terminal stops ) can be any combination of canonical stop codons ( TAA , TAG , TGA ) .
    \end{itemize}
    \item (SciSpacy)
    \begin{itemize}
        \item Two stop codons of an iORF ( i.e. the inframe and C - terminal stops ) can be any combination of canonical stop codons ( TAA , TAG , TGA ) .
    \end{itemize}
    \item (UDPIPE)
    \begin{itemize}
        \item Two stop codons of an iORF ( i.e. the inframe and C- terminal stops ) can be any combination of canonical stop codons ( TAA , TAG , TGA ) .
    \end{itemize}
\end{itemize}

\subsubsection{Words with letters and numbers}

\begin{displayquote}
\textbf{Example 12: “Selenocysteine and pyrrolysine are the 21st and 22nd amino acids, which are genetically encoded by stop codons.”}
\end{displayquote}

 \begin{itemize}
    \item (NLTK-space)
    \begin{itemize}
        \item Selenocysteine and pyrrolysine are the 21st and 22nd amino acids, which are genetically encoded by stop codons.
    \end{itemize}
    \item (NLTK-tb, spacy, SciSpacy, Stanza, StanzaCraft, Tokenizers, UDPIPE)
    \begin{itemize}
        \item Selenocysteine and pyrrolysine are the 21st and 22nd amino acids , which are genetically encoded by stop codons .
    \end{itemize}
\end{itemize}

\subsubsection{Words with numbers and one type of punctuation}

\begin{displayquote}
\textbf{Example 13: “A total of 26,003 iORF satisfied the above criteria.”}
\end{displayquote}

 \begin{itemize}
    \item (NLTK-space)
    \begin{itemize}
        \item A total of 26,003 iORF satisfied the above criteria.
    \end{itemize}
    \item(NLTK-tb, spacy, SciSpacy, Stanza, StanzaCraft, UDPIPE)
    \begin{itemize}
        \item A total of 26,003 iORF satisfied the above criteria .
    \end{itemize}
    \item(Tokenizers)
    \begin{itemize}
        \item A total of 26 , 003 iORF satisfied the above criteria .
    \end{itemize}
\end{itemize}

\begin{displayquote}
\textbf{Example 14: “The patient had prior x-ray on 1/2 which demonstrated no pneumonia.” }
\end{displayquote}
	
 \begin{itemize}
    \item (NLTK-space)
    \begin{itemize}
        \item The patient had prior x-ray on 1/2 which demonstrated no pneumonia.
    \end{itemize}
    \item(NLTK-tb, spacy, Stanza, StanzaCraft, Tokenizers)
    \begin{itemize}
        \item The patient had prior x-ray on 1/2 which demonstrated no pneumonia .
    \end{itemize}
    \item(SciSpacy)
    \begin{itemize}
        \item The patient had prior x - ray on 1/2 which demonstrated no pneumonia . 
    \end{itemize}
    \item(UDPIPE)
    \begin{itemize}
        \item The patient had prior x - ray on 1 / 2 which demonstrated no pneumonia .  
    \end{itemize}
\end{itemize}

\begin{displayquote}
\textbf{Example 15: “Indeed, it has been estimated recently that the current yeast and human protein interaction maps are only 50\% and 10\% complete, respectively 18.”}
\end{displayquote}

 \begin{itemize}
    \item (NLTK-space)
    \begin{itemize}
        \item Indeed, it has been estimated recently that the current yeast and human protein interaction maps are only 50\% and 10\% complete, respectively 18.
    \end{itemize}
    \item (NLTK-tb, spacy, SciSpacy, Stanza, StanzaCraft, Tokenizers, UDPIPE)
    \begin{itemize}
        \item Indeed , it has been estimated recently that the current yeast and human protein interaction maps are only 50 \% and 10 \% complete , respectively 18 .
    \end{itemize}
\end{itemize}

\begin{displayquote}
\textbf{Example 16: “The dotted line indicates significance level 0.05 after a correction for multiple testing.”}
\end{displayquote}

 \begin{itemize}
    \item (NLTK-space)
    \begin{itemize}
        \item The dotted line indicates significance level 0.05 after a correction for multiple testing.
    \end{itemize}
    \item (NLTK-tb, spacy, SciSpacy, Stanza, StanzaCraft, Tokenizers, UDPIPE)
    \begin{itemize}
        \item The dotted line indicates significance level 0.05 after a correction for multiple testing .
    \end{itemize}
\end{itemize}

\begin{displayquote}
\textbf{Example 17: “E-selectin is induced within 1-2 h, peaks at 4-6 h, and gradually returns to basal level by 24 h.”}
\end{displayquote}

\begin{itemize}
    \item (NLTK-space)
    \begin{itemize}
        \item E-selectin is induced within 1-2 h, peaks at 4-6 h, and gradually returns to basal level by 24 h.
    \end{itemize}
    \item (NLTK-tb)
    \begin{itemize}
        \item E-selectin is induced within 1-2 h , peaks at 4-6 h , and gradually returns to basal level by 24 h . 
    \end{itemize}
    \item (spacy)
    \begin{itemize}
        \item E-selectin is induced within 1 - 2 h , peaks at 4 - 6 h , and gradually returns to basal level by 24 h.
    \end{itemize}
    \item (SciSpacy)
    \begin{itemize}
        \item E - selectin is induced within 1 - 2 h , peaks at 4 - 6 h , and gradually returns to basal level by 24 h. 
    \end{itemize}
    \item (Stanza, StanzaCraft)
    \begin{itemize}
        \item E-selectin is induced within 1 - 2 h , peaks at 4 - 6 h , and gradually returns to basal level by 24 h .
    \end{itemize}
    \item (Tokenizers)
    \begin{itemize}
        \item E-selectin is induced within 1-2 h , peaks at 4-6 h , and gradually returns to basal level by 24 h .
    \end{itemize}
    \item (UDPIPE)
    \begin{itemize}
        \item E - selectin is induced within 1 - 2 h , peaks at 4 - 6 h , and gradually returns to basal level by 24 h . 
    \end{itemize}
\end{itemize}

\subsubsection{Numeration}

\begin{displayquote}
\textbf{Example 18: “1. Bioactivation of sulphamethoxazole (SMX) to chemically-reactive metabolites and subsequent protein conjugation is thought to be involved in SMX hypersensitivity.”}
\end{displayquote}

 \begin{itemize}
    \item (NLTK-space)
    \begin{itemize}
        \item 1. Bioactivation of sulphamethoxazole (SMX) to chemically-reactive metabolites and subsequent protein conjugation is thought to be involved in SMX hypersensitivity. 
    \end{itemize}
    \item (NLTK-tb, Tokenizers)
    \begin{itemize}
        \item 1. Bioactivation of sulphamethoxazole ( SMX ) to chemically-reactive metabolites and subsequent protein conjugation is thought to be involved in SMX hypersensitivity .
    \end{itemize}
    \item (spacy)
    \begin{itemize}
        \item 1 . Bioactivation of sulphamethoxazole ( SMX ) to chemically-reactive metabolites and subsequent protein conjugation is thought to be involved in SMX hypersensitivity . 
    \end{itemize}
    \item (SciSpacy, Stanza, StanzaCraft, UDPIPE)
    \begin{itemize}
        \item 1 . Bioactivation of sulphamethoxazole ( SMX ) to chemically - reactive metabolites and subsequent protein conjugation is thought to be involved in SMX hypersensitivity . 
    \end{itemize}
\end{itemize}

\subsubsection{Hypertext markup}

\begin{displayquote}
\textbf{Example 19: “Bcd mRNA transcripts of \&lt; or $=$ 2.6kb were selectively expressed in PBL and testis of healthy individuals.”}
\end{displayquote}

 \begin{itemize}
    \item (NLTK-space)
    \begin{itemize}
        \item Bcd mRNA transcripts of \&lt  or $=$ 2.6kb were selectively expressed in PBL and testis of healthy individuals. 
    \end{itemize}
    \item (NLTK-tb, Tokenizers)
    \begin{itemize}
        \item Bcd mRNA transcripts of \& lt   or $=$ 2.6kb were selectively expressed in PBL and testis of healthy individuals .
    \end{itemize}
    \item Spacy, SciSpacy, UDPIPE)
    \begin{itemize}
        \item Bcd mRNA transcripts of \& lt   or $=$ 2.6 kb were selectively expressed in PBL and testis of healthy individuals .
    \end{itemize}
    \item (Stanza, StanzaCraft)
    \begin{itemize}
        \item Bcd mRNA transcripts of \&lt   or $=$ 2.6 kb were selectively expressed in PBL and testis of healthy individuals .
    \end{itemize}
\end{itemize}

\subsubsection{URLs}

\begin{displayquote}
\textbf{Example 20: “Names of all available Trace Databases were taken from a list of databases at http://www.ncbi.nlm.nih.gov/blast/mmtrace.shtml”}
\end{displayquote}

 \begin{itemize}
    \item (NLTK-space, spacy, SciSpacy, Stanza, StanzaCraft, UDPIPE)
    \begin{itemize}
        \item Names of all available Trace Databases were taken from a list of databases at http://www.ncbi.nlm.nih.gov/blast/mmtrace.shtml
    \end{itemize}
    \item (NLTK-tb, Tokenizers)
    \begin{itemize}
        \item Names of all available Trace Databases were taken from a list of databases at http : //www.ncbi.nlm.nih.gov/blast/mmtrace.shtml
    \end{itemize}
\end{itemize}

\subsubsection{DNA Sequences}

\begin{displayquote}
\textbf{Example 21: “Footprinting analysis revealed that the identical sequence CCGAAACTGAAAAGG, designated E6, was protected by nuclear extracts from B cells, T cells, or HeLa cells.”}
\end{displayquote}

 \begin{itemize}
    \item (NLTK-space)
    \begin{itemize}
        \item Footprinting analysis revealed that the identical sequence CCGAAACTGAAAAGG, designated E6, was protected by nuclear extracts from B cells, T cells, or HeLa cells.
    \end{itemize}
    \item (NLTK-tb, spacy, SciSpacy, Stanza, StanzaCraft, Tokenizers, UDPIPE)
    \begin{itemize}
        \item Footprinting analysis revealed that the identical sequence CCGAAACTGAAAAGG , designated E6 , was protected by nuclear extracts from B cells , T cells , or HeLa cells .
    \end{itemize}
\end{itemize}

\subsubsection{Temporal Expressions}

\begin{displayquote}
\textbf{Example 22: “This was last documented on the Nuclear Cystogram dated 1/2/01.”}
\end{displayquote}

 \begin{itemize}
    \item (NLTK-space)
    \begin{itemize}
        \item This was last documented on the Nuclear Cystogram dated 1/2/01. 
    \end{itemize}
    \item (NLTK-tb, spacy, SciSpacy, Stanza, StanzaCraft, Tokenizers, UDPIPE)
    \begin{itemize}
        \item This was last documented on the Nuclear Cystogram dated 1/2/01 . 
    \end{itemize}
\end{itemize}

\subsubsection{Chemical Substances}

\begin{displayquote}
\textbf{Example 23: “These results reveal a central role for CaMKIV/Gr as a Ca(2+)-regulated activator of gene transcription in T lymphocytes.”}
\end{displayquote}

\begin{itemize}
    \item (NLTK-space)
    \begin{itemize}
        \item These results reveal a central role for CaMKIV/Gr as a Ca(2+)-regulated activator of gene transcription in T lymphocytes.
    \end{itemize}
    \item (NLTK-tb, Tokenizers)
    \begin{itemize}
        \item These results reveal a central role for CaMKIV/Gr as a Ca ( 2+ ) -regulated activator of gene transcription in T lymphocytes . 
    \end{itemize}
    \item (spacy)
    \begin{itemize}
        \item These results reveal a central role for CaMKIV/Gr as a Ca(2+)-regulated activator of gene transcription in T lymphocytes .
    \end{itemize}
    \item (SciSpacy)
    \begin{itemize}
        \item These results reveal a central role for CaMKIV / Gr as a Ca(2+)-regulated activator of gene transcription in T lymphocytes . 
    \end{itemize}
    \item (Stanza)
    \begin{itemize}
        \item These results reveal a central role for CaMKIV / Gr as a Ca ( 2 + ) - regulated activator of gene transcription in T lymphocytes .
    \end{itemize}
    \item (StanzaCraft)
    \begin{itemize}
        \item These results reveal a central role for CaMKIV / Gr as a Ca( 2 + ) - regulated activator of gene transcription in T lymphocytes .
    \end{itemize}
    \item (UDPIPE)
    \begin{itemize}
        \item These results reveal a central role for CaMKIV / Gr as a Ca ( 2+ ) - regulated activator of gene transcription in T lymphocytes . 
    \end{itemize}
\end{itemize}

\begin{displayquote}
\textbf{Example 24: “Expression of a highly specific protein inhibitor for cyclic AMP-dependent protein kinases in interleukin-1 (IL-1)-responsive cells blocked IL-1-induced gene transcription that was driven by the kappa immunoglobulin enhancer or the human immunodeficiency virus long terminal repeat.”}
\end{displayquote}

 \begin{itemize}
    \item (NLTK-space)
    \begin{itemize}
        \item Expression of a highly specific protein inhibitor for cyclic AMP-dependent protein kinases in interleukin-1 (IL-1)-responsive cells blocked IL-1-induced gene transcription that was driven by the kappa immunoglobulin enhancer or the human immunodeficiency virus long terminal repeat.
    \end{itemize}
    \item (NLTK-tb, Tokenizers)
    \begin{itemize}
        \item Expression of a highly specific protein inhibitor for cyclic AMP-dependent protein kinases in interleukin-1 ( IL-1 ) -responsive cells blocked IL-1-induced gene transcription that was driven by the kappa immunoglobulin enhancer or the human immunodeficiency virus long terminal repeat .
    \end{itemize}
    \item (spacy)
    \begin{itemize}
        \item Expression of a highly specific protein inhibitor for cyclic AMP-dependent protein kinases in interleukin-1 (IL-1)-responsive cells blocked IL-1-induced gene transcription that was driven by the kappa immunoglobulin enhancer or the human immunodeficiency virus long terminal repeat .
    \end{itemize}
    \item (SciSpacy)
    \begin{itemize}
        \item Expression of a highly specific protein inhibitor for cyclic AMP - dependent protein kinases in interleukin-1 ( IL-1)-responsive cells blocked IL-1 - induced gene transcription that was driven by the kappa immunoglobulin enhancer or the human immunodeficiency virus long terminal repeat .
    \end{itemize}
    \item (Stanza)
    \begin{itemize}
        \item Expression of a highly specific protein inhibitor for cyclic AMP - dependent protein kinases in interleukin - 1 ( IL - 1 ) - responsive cells blocked IL - 1 - induced gene transcription that was driven by the kappa immunoglobulin enhancer or the human immunodeficiency virus long terminal repeat . 
    \end{itemize}
    \item (StanzaCraft)
    \begin{itemize}
        \item Expression of a highly specific protein inhibitor for cyclic AMP -dependent protein kinases in interleukin-1 ( IL - 1 ) - responsive cells blocked IL - 1- induced gene transcription that was driven by the kappa immunoglobulin enhancer or the human immunodeficiency virus long terminal repeat . 
    \end{itemize}
    \item (UDPIPE)
    \begin{itemize}
        \item Expression of a highly specific protein inhibitor for cyclic AMP - dependent protein kinases in interleukin -1 ( IL - 1 ) - responsive cells blocked IL - 1 -induced gene transcription that was driven by the kappa immunoglobulin enhancer or the human immunodeficiency virus long terminal repeat .
    \end{itemize}
\end{itemize}

\subsection{Evaluating Variation in Tokenizer Outputs}

For each example, Table 2 reports the number of unique tokens and total tokens returned by each tokenizer; as well as the number of uniquely returned tokenization outputs. Table 3 investigates set agreement between returned tokenizer outputs.

The whitespace tokenizer performs differently than other tokenizers. The remaining seven tokenizers tend to cluster with respect to performance according to tokenizer methodology: i.e. rule-based systems versus neural classification systems. 

For none of the twenty-four example sentences included in our study, did all eight tokenizers agree on a single returned output. As expected, the whitespace tokenizer performed differently than the other tokenizers. Without post-hoc normalization heuristics, the white space tokenizer made no attempt to split tokens according to prefix/suffix/infix patterns, nor did it handle trailing punctuation affixed to tokens. Excluding the white space tokenizer, for eight of the twenty-four examples, the remaining 7 tokenizers agreed on a single output. This suggests that for certain sentences consisting of relatively simplistic token types, the tokenizers agree on an output tokenization. However, for certain sentences six (e.g. examples 4,10) or seven (e.g. examples 17, 23, 24) distinct outputs were returned by the tokenizers, suggesting certain challenging linguistic aspects associated with those particular sentences. Below we highlight excerpts from the example sentences where tokenizer variation was most pronounced. Tokens within these examples tend to be a mix of alphabetic/numeric characters, contain alphabetic characters in both lowercase/uppercase, and often include complex punctuation patterns as token infixes/suffixes.

\begin{itemize}
    \item Example 4: “The maximal effect is observed at the IL-10 concentration of 20 U/ml”
    \item Example 10: “…the CD4- CD8-, CD4+ CD8+, CD4+ CD8-, and CD4- CD8+ cell populations.”
    \item Example 17: “E-selectin is induced within 1-2 h, peaks at 4-6 h, and gradually returns…”
    \item Example 23: “…central role for CaMKIV/Gr as a Ca(2+)-regulated activator…”
    \item Example 23: “…interleukin-1 (IL-1)-responsive cells blocked IL-1-induced gene transcription…”
\end{itemize}

Considering some of the most challenging words/tokens identified in the examples above, we note that the white-space tokenizer (depending on purpose of study), may perform reasonably well. Other tokenizers varied in how they parsed the complex patterns of internal/trailing punctuation symbols embedded within tokens. It should be noted that the complex words/tokens also are the ones which convey the most precise meaning regarding linguistic information embedded in the document collection (hence, decisions regarding the operationalization of these symbols are crucially important).

\begin{table}[H]
	\caption{Number of distinct outputs returned by the eight tokenizers applied to each of the twenty-four example biomedical sentences from \citet{diaz2015analysis}. Number of unique/total tokens returned by the tokenizers (for each of the twenty-four examples); as well as accumulated over all examples in the twenty-four document corpus.}
	\centering
    \scalebox{0.75}{
	\begin{tabular}{lccccccccc}
		\toprule
	& Unique Outputs & nltk-space &	nltk-tb & spacy & scispacy & stanza & stanza-craft &
 tokenizers & udpipe \\
        &   & (unique/total) & (unique/total) & (unique/total) & (unique/total) & (unique/total) & (unique/total) & (unique/total) & (unique/total) \\
\midrule
Example 1	& 3   & 3/3   &	4/4   & 4/4   &	6/6   &	4/4   & 4/4   & 4/4   & 6/6 \\
\midrule
Example 2	& 3	  & 6/6	  & 7/7	  & 7/7   &	9/11  &	9/11  &	9/11  &	7/7   &	9/11 \\
\midrule
Example 3	& 3	  & 12/14 &	13/15 &	13/15 &	13/15 &	15/17 &	15/17 &	13/15 &	15/17 \\
\midrule
Example 4	& 6   &	12/12 &	13/13 &	13/13 &	15/15 &	16/16 &	15/15 &	13/13 &	17/17 \\
\midrule
Example 5	& 4	  & 26/29 &	27/30 &	27/30 &	31/34 &	31/34 &	31/34 &	27/30 &	32/35 \\
\midrule
Example 6	& 3	  & 23/26 &	24/27 &	24/27 &	24/27 &	24/27 &	24/27 &	24/27 &	25/28 \\
\midrule
Example 7	& 2   &	15/15 &	18/19 &	18/19 &	18/19 &	18/19 &	18/19 &	18/19 &	18/19 \\
\midrule
Example 8	& 2	  & 27/27 &	33/36 &	33/36 &	33/36 &	33/36 &	33/36 &	33/36 &	33/36 \\
\midrule
Example 9	& 3	  & 22/23 &	25/27 &	25/27 &	25/27 &	25/27 &	25/27 &	25/27 &	26/28 \\
\midrule
Example 10	& 6	  & 15/18 &	16/22 &	17/26 &	17/26 &	16/30 &	19/27 &	16/22 &	17/29 \\
\midrule
Example 11	& 4  &	20/23 &	24/30 &	24/30 &	26/32 &	24/30 &	24/30 &	24/30 &	25/31 \\
\midrule
Example 12	& 2  &	15/17 &	17/19 &	17/19 &	17/19 &	17/19 &	17/19 &	17/19 &	17/19 \\
\midrule
Example 13	& 3  &	9/9  & 10/10  &	10/10 &	10/10 &	10/10 &	10/10 &	12/12 &	10/10 \\
\midrule
Example 14  & 4  & 11/11 & 12/12  & 12/12 & 14/14 & 12/12 & 12/12 & 12/12 & 16/16 \\
\midrule
Example 15 & 2  & 22/23 & 25/28 & 25/28 & 25/28 & 25/28 & 25/28 & 25/28 & 25/28 \\
\midrule
Example 16 & 2 & 13/13  & 14/14 & 14/14 & 14/14 & 14/14 & 14/14 & 14/14 & 14/14 \\
\midrule
Example 17 & 7 & 18/19  & 19/22 & 22/25 & 23/27 & 22/26 & 22/26 & 19/22 & 23/28 \\
\midrule
Example 18 & 4  & 19/20 & 21/23 & 21/24 & 23/26 & 23/26 & 23/26 & 21/23 & 23/26 \\
\midrule
Example 19 & 4  & 18/19 & 20/22 & 21/23 & 21/23 & 20/22 & 20/22 & 20/22 & 21/23 \\
\midrule
Example 20 & 2  & 14/15 & 16/17 & 14/15 & 14/15 & 14/15 & 14/15 & 16/17 & 14/15 \\
\midrule
Example 21 & 2  & 22/23 & 23/28 & 23/28 & 23/28 & 23/28 & 23/28 & 23/28 & 23/28 \\
\midrule
Example 22 & 2  & 10/10 & 11/11 & 11/11 & 11/11 & 11/11 & 11/11 & 11/11 & 11/11 \\
\midrule
Example 23 & 7 & 17/18 & 22/23 & 18/19 & 20/21 & 26/27 & 25/26 & 22/23 & 25/26 \\
\midrule
Example 24 & 7 & 34/36 & 38/40 & 35/37 & 39/42 & 41/51 & 42/47 & 38/40 & 42/49 \\
\midrule
Total Corpus & --- & 289/429 & 294/499 & 294/499 & 303/526 & 298/540 & 300/531 & 294/501 &	306/550 \\
        \bottomrule
	\end{tabular}
	}
	\label{tab:table2}
\end{table}

 \begin{table}[H]
	\caption{Jaccard index quantifying the magnitude of similarity of outputs returned from each of the eight tokenizers applied to the twenty-four problematic and challenging biomedical sentences outlined in \citet{diaz2015analysis}.}
	\centering
    \scalebox{0.925}{
	\begin{tabular}{lcccccccc}
		\toprule

 & nltk-space & nltk-tb & spacy & scispacy & stanza & stanza-craft & tokenizers & udpipe \\
\midrule
nltk-space	& 1.000	& 0.690	& 0.680	& 0.609	& 0.613	& 0.623	& 0.685	& 0.570 \\
\midrule
nltk-tb 	& 0.690 & 1.000	& 0.915	& 0.826	& 0.822	& 0.833	& 0.993 & 0.786 \\
\midrule
spacy	    & 0.680 & 0.915	& 1.000	& 0.889	& 0.873	& 0.886	& 0.909	& 0.829 \\
\midrule
scispacy	& 0.609	& 0.826	& 0.889	& 1.000	& 0.920	& 0.908	& 0.820	& 0.909 \\
\midrule
stanza	    & 0.613 & 0.822	& 0.873	& 0.920	& 1.000 & 0.954	& 0.816	& 0.917 \\
\midrule
stanza-craft & 0.623 & 0.833 & 0.886 & 0.908 & 0.954 & 1.000 & 0.828 & 0.888 \\
\midrule
tokenizers	& 0.685	& 0.993	& 0.909	& 0.820 & 0.816	& 0.828	& 1.000 & 0.780 \\
\midrule
udpipe	     & 0.570 & 0.786 & 0.829 & 0.909 & 0.917 & 0.888 & 0.780 & 1.000 \\
        \bottomrule
	\end{tabular}
	}
	\label{tab:table3}
\end{table}

\section{Discussion}

This study observed variation in returned outputs, when comparing different tokenization algorithms applied across a set of twenty-four problematic and challenging biomedical sentences. \citet{diaz2015analysis} noted similar variation when comparing twelve tokenizers across these same example sentences. Similar issues regarding variation in tokenizers applied to biomedical text has been discussed in \citet{he2006comparison} and \citet{jiang2007empirical}.

We observe that tokenizers relying on similar underlying computational/statistical methods tended to cluster in their performance (Table 3). The white-space tokenizer performed differently from the other tokenizers. The tokenizers implementing rule-based systems, using regular expressions and/or pattern matching performed similarly (e.g. nltk-tb, spacy, scispacy, tokenizers). The tokenizers using neural networks for tagging sequential data to classify token boundaries performed similarly (e.g. stanza, stanza-craft, udpipe). Compared to the twelve tokenizers outlined in \citet{diaz2015analysis}, the eight general purpose Python/R tokenizers showed slightly increased variation in returned outputs. In general, the tokenized outputs returned in \citet{diaz2015analysis} were a subset of the outputs returned in our study. Again, the increased variation is largely a result of how the general purpose Python/R tokenizers handled tokenization of complex biomedical words, characterized by punctuation infixes/suffixes. That said, it is encouraging that general purpose tokenizers broadly available in R/Python resulted in similar tokenization outputs as bespoke clinical tokenizers investigated in \citet{diaz2015analysis}, which are not as accessible to data scientists who often rely on Python/R languages for statistical computing. 

Using a simple whitespace tokenizer as a baseline exaggerated differences between tokenization routines. In practice, post-hoc normalization algorithms could be applied to white-space tokenized outputs, to obtain a token-set satisfactory for a particular research purpose. Certain normalization steps which could be applied include: removal of punctuation/numbers from tokens, case-folding (converting all alphabetic characters to a single case), removal of stop-words, removal of short words/tokens (e.g. single character tokens), removal of words based on corpus occurrence frequencies, stemming or lemmatization, spell-checking and acronym expansion. 

We chose to focus on eight modern (at the time of writing), and open-source tokenizers, available in popular data scientific languages (Python/R). \citet{diaz2015analysis} compare other tokenizers, perhaps with more focused suitability on biomedical/clinical documents (e.g. cTAKES, MetaMap, etc.). We have not explored tokenizers relevant to modern neural network frameworks for processing sequential data (for example, transformer models); which often employ character-level tokenizers or byte-level tokenizers. These tokenizers represent sub-word level tokens as a distinct type, and are associated with particular embedding vectors (which can be compositionally integrated to yield a single semantic representation vector). For example: PyTorch (torchtext), TensorFlow (TF text) or HuggingFace tokenizers (which may be specific to a given neural model).  

The evaluation of tokenization algorithms used in this study was descriptive. Different study designs have been used to evaluate tokenizer performance. Certain authors apply tokenizers across several different text corpora, and report variation in corpus level summary statistics. Other authors interested in downstream supervised machine learning objectives incorporate tokenizer evaluation as a discrete tunable hyper-parameter within a larger data scientific pipeline \citet{hacohen2020influence}. Less work has been conducted to understand the impact of text pre-processing on unsupervised machine learning models (e.g. document clustering, topic models, word embedding models, etc.). Our approach to evaluation is necessarily simplistic; however, is also quite revealing with regards to the types of words tokenizers find most challenging/problematic to consistently process. 

It may be possible to engage with statistical semantic models without engaging with tokenizers. For example, lexicon-based methodologies could be applied to many of these examples, where only a certain subset of words (or word groups) are extracted from digital character sequences, based on a priori defined subject matter knowledge (e.g. dictionaries/lexicons). Alternatively, one could attempt to extract counts of mentions related to certain biomedical concepts from sentences (represented as digital character sequences). For example, the quickUMLS program \citep{soldaini2016quickumls} would map inputs texts to a vector of UMLS CUI (concept unique identifier) counts. 

Tokenization performance should be fit for purpose. Hybrid systems employing both computational tokenization algorithms and human subject-matter knowledge may perform well for certain downstream tasks. For example, a hybrid system may involve an initial “tokenization pass” over the corpus; whereby, all inputs texts are tokenized, and corpus-level tokenization summary statistics are gathered. Next, a human would review the most frequently occurring tokens in the corpus, and may decide which elements of the token-list should be included/excluded. Choice of words to include/exclude is necessarily subjective (and ought to be dependent on the overarching aims of the research project). Included tokens could be further normalized, reducing lexical variation, and focusing on semantically related concept groups. The hybrid method trades off human-versus-computational time, intersecting with quality/interpretability.

\section{Conclusions}

We observe variation in tokenizer outputs, when comparing tokenization performance using several challenging biomedical example sentences. Words which were difficult to consistently tokenize included complex punctuation characters as suffixes/infixes. Several of the modern tokenizers being compared performed effectively when applied across several challenging biomedical sentences. However, subtle nuances exist in how difficult examples were tokenized. Data scientists engaging with text mining should be familiar with the landscape of tokenizers available for their particular research problem, and how the choice of tokenizer may impact downstream inferences.  

\newpage

\bibliographystyle{unsrtnat}
\bibliography{references}  

\clearpage

\end{document}